\documentclass[journal]{IEEEtran}
\ifCLASSINFOpdf
\else
\fi
\usepackage{amsfonts}
\usepackage{graphicx}
\usepackage{epstopdf}
\usepackage{amsmath}
\usepackage{amssymb}
\usepackage{algorithm}
\usepackage{algorithmicx}
\usepackage{algpseudocode}
\usepackage{multirow}

\begin{document}
%
\title{A Unified Framework for Biphasic Facial Age Translation with Noisy-Semantic Guided Generative Adversarial Networks
}
%
%
%

\author{Muyi Sun,
        Jian Wang,
        Qi Li,~\IEEEmembership{Member,~IEEE,}
        Yunfan Liu,
        Zhenan Sun,~\IEEEmembership{Senior Member,~IEEE,}
\thanks{Muyi Sun, Jian Wang, 
Qi Li, Yunfan Liu, Zhenan Sun are with the Center for Research on Intelligent Perception And Computing, National Laboratory of Pattern Recognition, Institute of Automation, Chinese Academy of Sciences, Beijing 100190, China. (e-mail: muyi.sun@cripac.ia.ac.cn; jian.wang@cripac.ia.ac.cn; 
qli@nlpr.ia.ac.cn; yunfan.liu@cripac.ia.ac.cn;   znsun@nlpr.ia.ac.cn).

\emph{\textbf{Corresponding author: Zhenan Sun. (znsun@nlpr.ia.ac.cn)}}}}

%
%

\markboth{IEEE TIFS Files, ~Vol.~x, No.~x, xxxx~xxxx}%
{Sun\MakeLowercase{\textit{et al.}}: Biphasic Facial Age Translation}
%



\maketitle

\begin{abstract}
Biphasic facial age translation aims at predicting the appearance of the input face at any age.
Facial age translation has received considerable research attention in the last decade due to its practical value in cross-age face recognition and various entertainment applications.
However, most existing methods model age changes between holistic images, regardless of the human face structure and the age-changing patterns of individual facial components.
Consequently, the lack of semantic supervision will cause infidelity of generated faces in detail.
To this end, we propose a unified framework for biphasic facial age translation with noisy-semantic guided generative adversarial networks.
Structurally, we project the class-aware noisy semantic layouts to ``soft" latent maps for the following injection operation on the individual facial parts.
In particular, we introduce two sub-networks, ProjectionNet and ConstraintNet.
ProjectionNet introduces the low-level structural semantic information with noise map and produces ``soft" latent maps.
ConstraintNet disentangles the high-level spatial features to constrain the ``soft" latent maps, which endows more age-related context into the ``soft" latent maps.
Specifically, attention mechanism is employed in ConstraintNet for feature disentanglement.
Meanwhile, in order to mine the strongest mapping ability of the network, we embed two types of learning strategies in the training procedure, supervised self-driven generation and unsupervised condition-driven cycle-consistent generation.
As a result, extensive experiments conducted on MORPH and CACD datasets demonstrate the prominent ability of our proposed method which achieves state-of-the-art performance.

\end{abstract}


%
\IEEEpeerreviewmaketitle

\section{Introduction}
\IEEEPARstart{B}{iphasic} facial age translation, also known as face progression and regression, aims at rendering age-changing effects on the input face image at any age (or age group).
High-fidelity face age generation can arouse people's good memories of their youth and beautiful expectations for the future.
Due to the broad applications of facial age translation in various domains such as cross-age face recognition and digital entertainment, a great number of studies have been put forward in the last two decades \cite{1}-\cite{3}.
However, due to the lack of paired age datasets and the intrinsic complexity of age translation task (involving changes in both shape and appearance of different facial components), age progression and regression is still a challenging problem.
Meanwhile, limited by the learning abilities and computing costs of the current algorithms, age translation tasks are mostly implemented through single-direction generation between single age pair (or age group pair) \cite{4}-\cite{6}, which will bring huge complexity to the training of the model among all ages (or age groups).
Therefore, a \textbf{Unified High-fidelity Biphasic} age translation method is indispensable. 
The illustration of biphasic face progression and regression is shown in Fig. 1. 

\begin{figure}[t]
\centering
\includegraphics[width=0.95\linewidth]{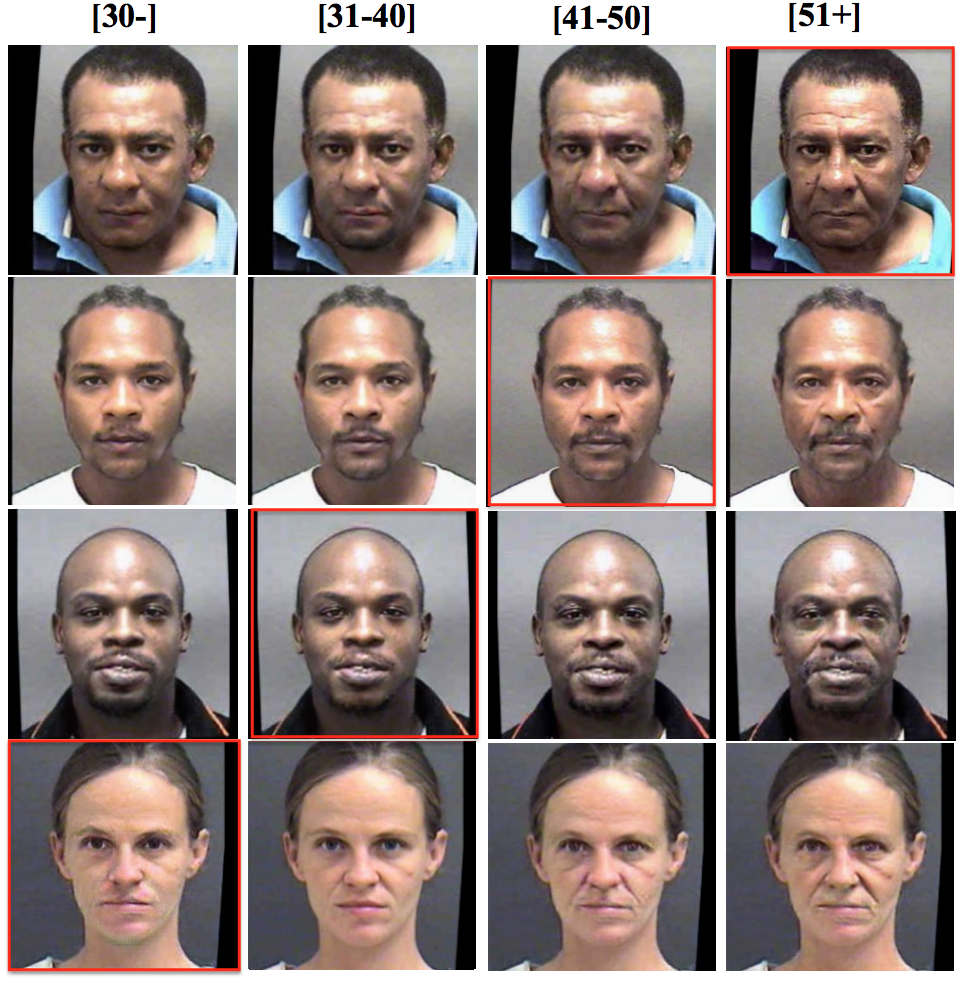}
\caption{Examples of biphasic facial age translation. For each row, the input image is framed by a red rectangular box, and rest images demonstrate the predicted appearances of the subject at three different age groups.
For each input images, face progression and regression can be achieved at the same time in a unified framework.}
\label{fig1}
\end{figure}

The study of age progression and regression could be dated back to the 1980s, where researches mainly relied on \textbf{model-based methods} to simulate the profile growth of human faces.
Todd et al. \cite{7} have modeled the change of facial appearance about time by a variant of cardioidal-strain transformation.
Lanitis et al. \cite{6} and Wang et al. \cite{8} have proposed structural strategies to study the problem from other biological aspects such as muscles and the structure of facial components.
However, these methods are computationally expensive and difficult to generalize high-fidelity faces, since detailed empirical aging rules are used to build the models.

With the increasing amount of available face age data, \textbf{prototype-based methods} have been developed.
As one of the pioneering works, Burt et al. \cite{9} have divided the face images into several groups according to age annotations.
Then, the average face (i.e., prototype) has been computed for each age group.
The mappings between prototypes are considered as age transformations for corresponding age groups.
Based on this scheme, more approaches have been proposed to improve the quality of generated face images \cite{10}-\cite{13} with various prior knowledge.
However, computing average faces would blur the image content heavily and thus could damage both the texture related to age changes and the personalized identity information.

Inspired by the remarkable development of Generative Adversarial Networks (GANs) \cite{14}, most recent works have resorted to conditional \textbf{GAN-based methods} to solve the problem of age progression and regression. 
Significant improvements have been achieved in the realism of generation results \cite{2}, \cite{4}-\cite{5}, \cite{15}-\cite{17}.
Zhang et al. \cite{2} have proposed a manifold learning approach to model the identity and age changes simultaneously.
Yang et al. \cite{4} have employed a pre-trained age estimation network in discriminator to enhance age-related features.
Tang et al. \cite{15} and Liu et al. \cite{5} have introduced identity consistency supervisions and wavelet-domain constraint for improving fidelity respectively.
However, these methods model age changes as mappings between images of holistic faces and propose various global constraints to regulate the training process.
They lost the control of the details in different semantic components, since the aging patterns for different facial components have little in common in terms of both geometry and appearance.
Therefore, Li et al. \cite{16} have proposed a patch-based method. Liu et al.\cite{17} has introduced the spatial attention mechanism to restrict manipulations on age-related regions.
Although some effects have been improved, their results still show blurred areas and artifacts.

To tackle the above-mentioned issues, we propose a unified framework for biphasic facial age translation with noisy-semantic guided generative adversarial networks.
Different from existing methods which focus on global image generation, we project the class-aware noisy semantic layouts to a group of ``soft" latent maps and then inject these maps into the generator to guide detailed age translation on the individual facial parts.
Specifically speaking, we introduce two sub-networks to explore low-level structural information and high-level features, which are ProjectionNet and ConstraintNet.
ProjectionNet combines the low-level structural semantic layouts with the ``soft" age condition and projects them to ``soft" latent maps. 
ConstraintNet disentangles the high-level spatial features to constrain the ``soft" latent maps, which endows more  age-related context into the ``soft" latent maps.
We employ the attention mechanism in the ConstraintNet for feature disentanglement.
For further promoting the training procedure, we also embed two types of learning strategies, supervised self-driven generation and unsupervised condition-driven cycle-consistent generation.

The main contributions are summarized as follows:
\begin{itemize}
\item A unified framework is proposed for biphasic facial age translation with noisy-semantic guided generative adversarial networks. 
Noisy-Semantic layouts are injected into the generator as the guidance of age transformation.  
\item Two sub-networks, ProjectionNet and ConstraintNet are proposed to introduce low-level structural semantic information and high-level spatial features for fine-grained high-fidelity age translation before latent map injection.
Attention mechanism is used in ConstraintNet for feature disentanglement.
\item Two types of learning strategies are jointly employed to promote the training procedure, supervised self-driven generation and unsupervised condition-driven cycle-consistent generation.
\item Extensive experiments are conducted on MORPH and CACD datasets.
Both qualitative and quantitative results demonstrate the effectiveness of the proposed method which achieves state-of-the-art performance. 
\end{itemize}

The rest of the paper is organized as follows: 
Section II presents a brief description of the related works. 
Section III introduces our proposed network architecture in detail. 
Section IV introduces the face datasets, evaluation metrics used in this paper and describes the implementation details. 
Meanwhile, the experimental results and ablation analysis are reported. 
Finally, in Section V, we conclude this work and further give some detailed discussions about age synthesis.

\section{Related Works}
In this section, we present a brief literature review on the several main fields related to our method, which are facial age synthesis, semantic driven image translation, attention-based feature fusion and disentanglement. 

\begin{figure*}[t]
\centering
\includegraphics[width=1.00\linewidth]{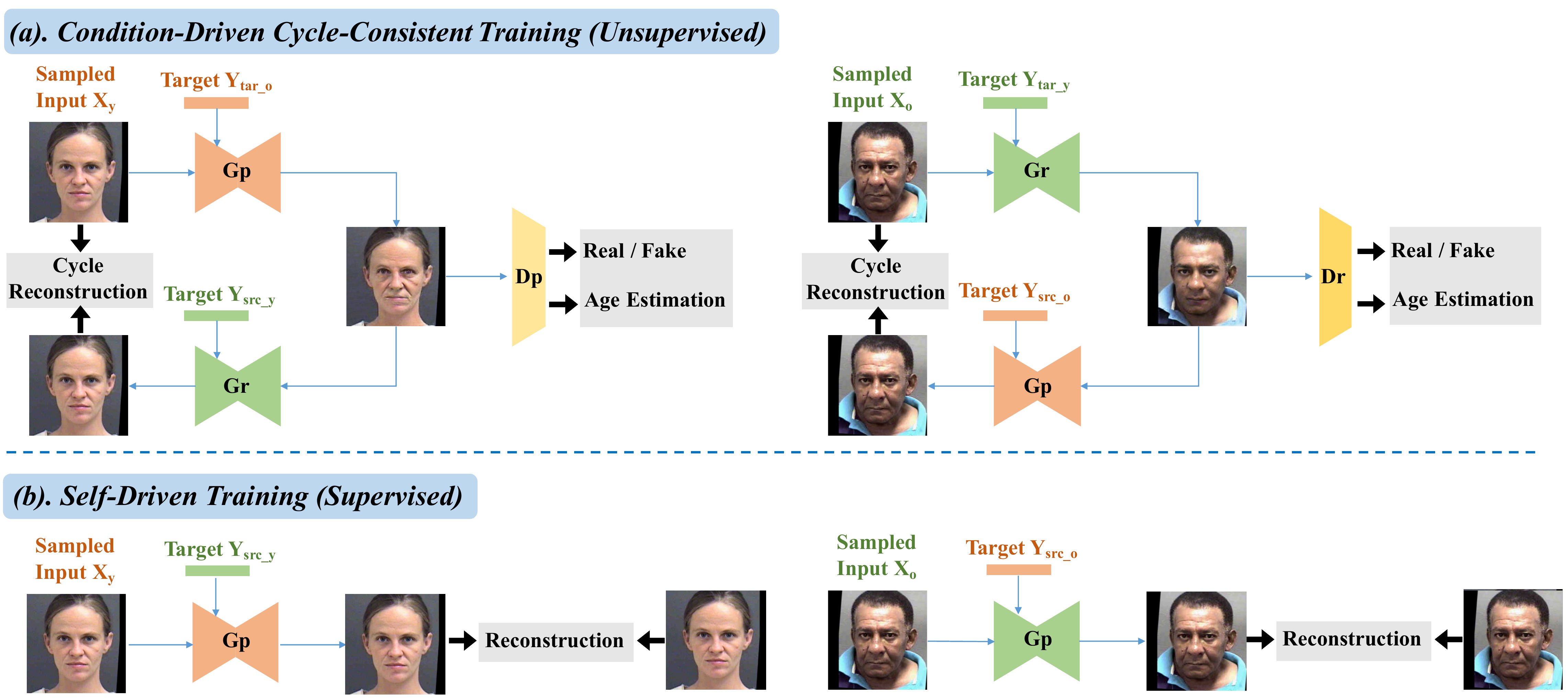}
\caption{The conditional cycle-consistent framework and the two types of learning strategies jointly used in this paper.
(a).unsupervised condition-driven cycle-consistent generation.
(b). supervised self-driven generation.
For each training iteration, we randomly sample a batch of face pairs (e.g. $\mathbf{x}_y$ and $\mathbf{x}_o$ ) to jointly implement $\mathbf{x}_y$ $\rightarrow$ $\mathbf{x}_{y\rightarrow o}$ $\rightarrow$ $\mathbf{x}_y'$, 
$\mathbf{x}_y$ $\rightarrow$ $\mathbf{x}_y'$  
and $\mathbf{x}_o$ $\rightarrow$ $\mathbf{x}_{o\rightarrow y}$ $\rightarrow$ $\mathbf{x}_o'$, 
$\mathbf{x}_o$ $\rightarrow$ $\mathbf{x}_o'$,
where each face pair has two face images with different ages.
We use this framework to train the two important generators $\mathbf{G}_p$ and $\mathbf{G}_r$ simultaneously.}
\label{fig:comparison}
\end{figure*}


\subsection{Facial Age Synthesis}
Facial age synthesis receives wide attention in recent years with the development of face forgery technologies.
Due to the lack of paired age data and the difficulty to obtain single-person face images with long duration, GAN-based methods make great progress on the face aging task in recent years \cite{15}-\cite{17}, acting as an unsupervised manner.
GAN-based methods could establish the mapping between the data distributions of two age groups.
In previous studies \cite{4}-\cite{5}, most GAN-based methods choose to model each age-pair in single direction (young to old, or old to young).
PAG-GAN \cite{4} has proposed to employ a network pre-trained on the age estimation task to extract age-related features in the discriminator.
Wavelet-GAN \cite{5} has introduced the facial attribute vector and a wavelet-domain constraint into the GANs to maintain the synthesized face attributes and obtain fidelity results.
However, these methods will bring huge computational cost.
For $n$ age groups, researchers need to build $n(n-1)$ models to settle the translations among all age groups. 
Although satisfactory results can be achieved in this way,  the balance between computing complexity and fidelity is still an urgent problem.
Furthermore, most methods \cite{15}-\cite{17} model age changes between holistic face images or face patches. 
They lost the control of the details in different semantic  components, since the age changes remain consistent in the respective semantic regions .
In this paper, we propose a conditional CycleGAN-based \cite{18} framework to jointly solve the translations among all age groups. 
And to improve the attention to different semantic regions, we design a specific semantic injection architecture, which utilizes the low-level structured information and high-level spatial features.

\subsection{Semantic Guided Image Synthesis}
Semantic guided image synthesis aims to convert an input segmentation mask to a photo-realistic image \cite{20}.
The advanced methods also use semantic layouts for image manipulation and editing \cite{21}-\cite{25}.
Pix2Pix \cite{20} first introduces conditional GANs into semantic driven image synthesis.
CycleGAN \cite{18} proposes a cycle consistency constraint for unpaired image translation including semantic image synthesis.
CoCosNet \cite{23} puts forward a cross-domain correspondence learning for semantic image translation.
In addition, Sesame \cite{24} utilizes local semantic map as the guidance to manipulate masked images.
SLMNet \cite{25} proposes a high-resolution sparse attention module that effectively transfers visual details to ‘new’ layouts at a resolution up to 512 × 512.
In the above methods, semantic layouts are directly used as the network input and transformed by convolutional representation.
Although their results are satisfactory, they still need to improve the details.

Different from the above, another part of methods are proposed to inject semantic layouts for image synthesis through redesign of normalization operations, which could enhance the depiction of details.
SPADE \cite{26} introduces a spatially-adaptive normalization for semantic-image style transfer.
SEAN \cite{27} updates SPADE and proposes to code region-wise features into respective feature vectors. 
Then the vectors are injected into the backbone network for different semantics.
CLADE \cite{28} proposes an class-adaptive normalization method for efficient semantic image synthesis.
Furthermore, Alharbi et al.\cite{29} put forward a structured noise injection method for disentangled face generation, which proves that noise could play an important role in the generation task.
Inspired by SPADE, CLADE and structured noise, in this paper, we apply the semantic injection into the generator and propose the noisy semantic for fine-grained generation, both firstly introduced for face age synthesis. 

\subsection{Attention based Feature Fusion and Disentanglement}
Attention mechanism has achieved great success in image generation.
For different purposes, attention modules are widely used for feature fusion and feature disentanglement.
Zhang et al. \cite{30} explore long-range dependencies and leverage self-attention mechanism for image generation.
Tang et al. \cite{21} propose a multi-channel attentive selection module to promote intermediate generation for high quality synthesis.
Diganta et al. \cite{32} capture fully dependencies across dimensions when computing attention weights.
Attention mechanism improves the utilization efficiency of feature information through these ‘re-weighted’ fusion strategy.

For attention-based feature disentanglement, 
Li et al. \cite{33} employ spatial attention to decompose the ID-related feature, and proposes an adaptive denormalization layer for cross-face ID swapping.
Pumarola et al. \cite{47} design a one-shot facial expression synthesis method based on attention mask.
Motivated by [47] and the Age-Invariant face recognition framework \cite{34}, we integrate the similar attention mechanism into two modules for face age synthesis, which are attention-based Feature Refinement Module (FRM) for feature fusion and the ConstraintNet for age-related feature decoupling.

\begin{figure*}[t]
\centering
\includegraphics[width=1.05\linewidth]{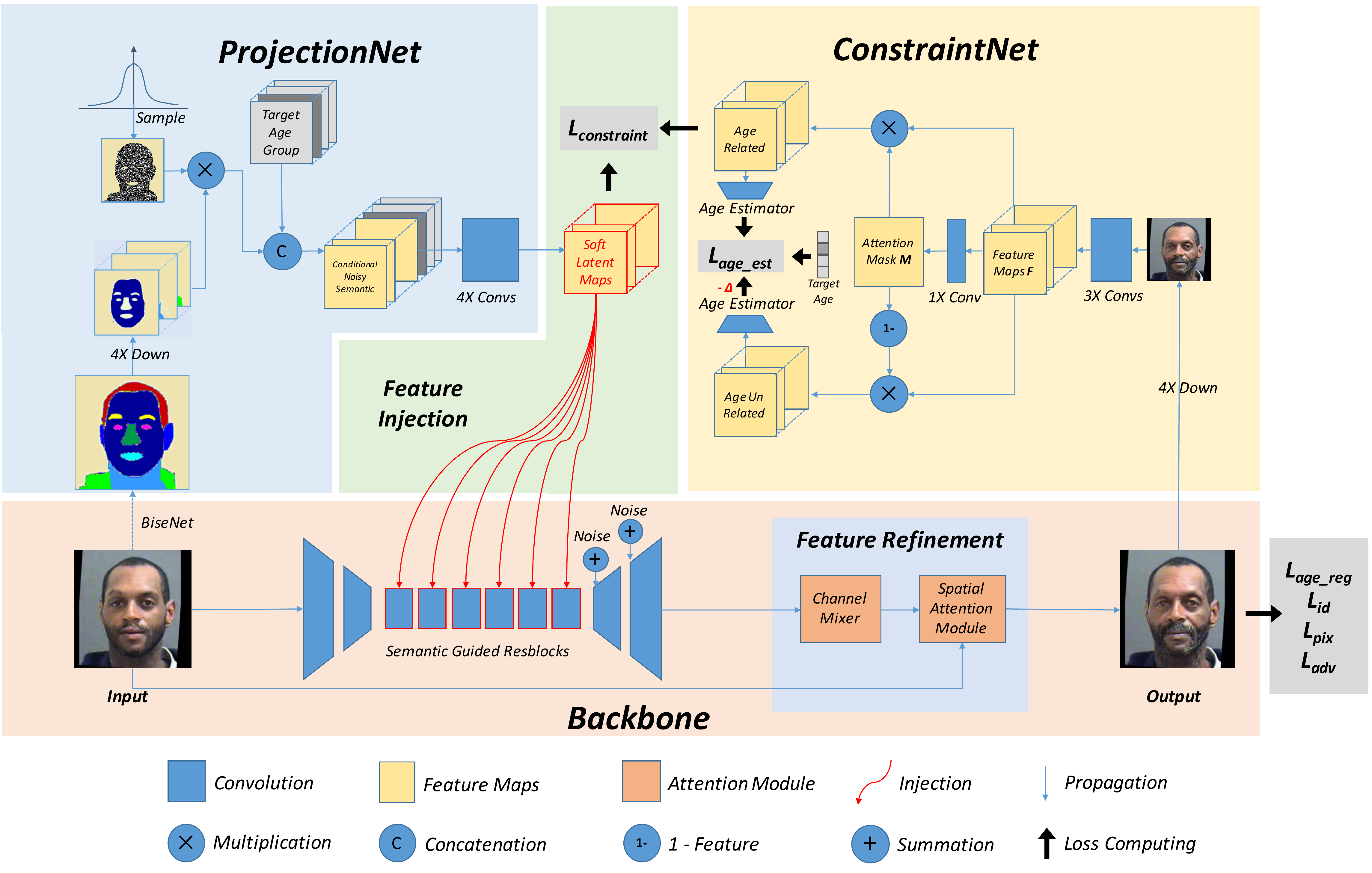}
\caption{The overall framework of our generator G.
The generator consists of two main parts, Noisy-Semantic Guided Encoder-decoder Architecture and the Attention-based Feature  Refinement Module.
The Noisy-Semantic Guided Module consists of three parts,  ProjectionNet, ConstraintNet and Latent Map Injection Module.
ProjectionNet combines the ``hard" age condition maps (e.g. 0-1-0-0) with noisy-semantic layouts and then projects them to  ``soft" latent maps.
ConstraintNet disentangles the age-related features and then  uses  these  features  to  constrain  the ``soft" latent maps, which endows the ``soft" latent maps with more age-related spatial context.}
\label{fig:comparison}
\end{figure*}

\section{The Proposed Method}
In this section, we firstly introduce the conditional cycle-consistent backbone and the jointly training strategy of our method.
Then, we describe the noisy-semantic guided Encoder-Decoder and the two corresponding embeded sub-networks, ProjectionNet and ConstraintNet. 
In addition, we analyze the application of the attention-based Feature Refinement Module in the generator.
Meanwhile, we review the spatially-adaptive normalization, feature disentanglement strategy, and illustrate their integration modes in our framework. 
Next, we introduce our overall framework and give a concrete illustration.
In the end, we give the objective functions and show the algorithm flow.
\subsection{Conditional Cycle-consistent Framework}
Due to the lack of paired data in age-related task, we choose conditional CycleGAN as our backbone, which could establish the mappings among all age groups. 
Noisy-semantic guided GANs performs biphasic facial age translation in two generation processes respectively.
Therefore, two generators ($G_p, G_r$) and two discriminators ($D_p, D_r$) are employed in the model as shown in Fig.2.
Given a young face image $\mathbf{x}_y$ at source young age $\mathbf{y}_{src_y}$ and given the target old age $\mathbf{y}_{tar_o}$ ($\mathbf{y}_{tar_o} > \mathbf{y}_{src_y}$), the age progressor $G_p$ is employed to produce the corresponding old face image $\mathbf{x}_{y\rightarrow o}=G_p(\mathbf{x}_y, \mathbf{y}_{tar_o})$.
To enforce cycle consistency, $\mathbf{x}_o$ is reversely mapped back to the age group $\mathbf{y}_{src}$ to reconstruct $\mathbf{x}_y$, i.e., $\mathbf{x}_y'=G_r(\mathbf{x}_{y\rightarrow o}, \mathbf{y}_{src_y})$.
The realism of generated face are supervised by $D_p$ and $D_r$ via adversarial learning  respectively.
Similarly, given an old face image $\mathbf{x}_o$ at source old age $\mathbf{y}_{src_o}$ and given the target young age $\mathbf{y}_{tar_y}$, the age translation model is the reverse of the above process.
For each training iteration, we sample a batch of face pairs to jointly implement $\mathbf{x}_y$ $\rightarrow$ $\mathbf{x}_{y\rightarrow o}$ $\rightarrow$ $\mathbf{x}_y'$  and $\mathbf{x}_o$ $\rightarrow$ $\mathbf{x}_{o\rightarrow y}$ $\rightarrow$ $\mathbf{x}_o'$, where each face pair has two face images with different ages. The conditional cycle-consistent framework is shown in Fig.2(a).

\subsection{Jointly Training Strategy}
Due to the inherent disadvantage of lacking paired data in the biphasic facial age translation task, previous methods mostly use original GAN based methods or CycleGAN based methods, which establish the mapping between two age groups once time.
In recent years, self-driven/self-supervised training methods have shown excellent performance in generation tasks \cite{34},\cite{38}.
In the scheme of self-driven training, the input appearance  and the target attribute condition are distributed into two decoupled branches.
If the target condition is set to be the same as the attribute of the input, the output of the network could be regarded as the reconstruction of the input, which could be supervised by itself.
In this way, we could introduce supervised learning into the unsupervised tasks.
We call this supervised self-driven learning strategy.
Then in the inference process, we could achieve conditional image translation by changing the target condition for different image groups/labels.  
In this paper, we jointly utilize two types of learning strategies to promote the training procedure, unsupervised training and supervised training, as shown in Fig.2.
In each iteration, we send a batch of two faces with different identities and ages into the networks to  implement 
$\mathbf{x}_y$ $\rightarrow$ $\mathbf{x}_{y\rightarrow o}$ $\rightarrow$ $\mathbf{x}_y'$,
$\mathbf{x}_o$ $\rightarrow$ $\mathbf{x}_{o\rightarrow y}$ $\rightarrow$ $\mathbf{x}_o'$
and 
$\mathbf{x}_y$ $\rightarrow$ $\mathbf{x}_y'$, 
$\mathbf{x}_o$ $\rightarrow$ $\mathbf{x}_o'$
concurrently.

\subsection{Generator}
In the framework, the generators $\mathbf{G}_p$ and $\mathbf{G}_r$ share the same structure.
Each generator consists of two parts, the Noisy Semantic Guided Encoder-decoder Architecture and the Attention-based Feature Refinement Module.
In the encoder-decoder, semantic injection is employed in the intermediate resblocks.
The illustration of the generator is shown in Fig.3.

\subsubsection{Noisy Semantic Guided Architecture}
We introduce the noisy semantic layouts into the age translation, aiming at promoting the network perception of individual facial components. 
The Noisy Semantic Guided Module consists of three parts, ProjectionNet, ConstraintNet and Latent Map Injection Module.

There are two core concepts in this module design.
\emph{\textbf{(1). Semantic Injection.}}
First of all, conditional information incorporated by concatenation would be inevitably washed off during the forward process \cite{26}, while layer-wise injection could ensure that parsing layouts and conditional information do participate in most steps of the generation process as illustrated in the backbone of Fig.3. 
Secondly, based on the observation that age changes are mostly represented by translation of facial textures (e.g., wrinkles and laugh lines), we attribute aging and rejuvenation as a special case of style transfer, and propose to involve the parsing maps as modulating coefficients with spatial information.
\emph{\textbf{(2). Noisy Semantic.}}
As described in CLADE [28], SPADE still has the inherent problem that large semantic regions on the generated images are smooth and blurred. There are also artifacts existing at the boundaries.
In age translation, these problems also emerge as shown in Fig.4.
Noise is proved to be effective to improve the randomness and fidelity in detail \cite{29}, \cite{48}.
In order to avoid region-wise smoothing, enhance the randomness of generation and refine the details, 
we introduce randomly sampled noisy maps with different resolutions into ProjectionNet and the decoder respectively, which could improve the visual fidelity and quantitative metrics simultaneously.
We call this type of layouts Noisy-Semantic.

\begin{figure}[t]
\centering
\includegraphics[width=1.0\linewidth]{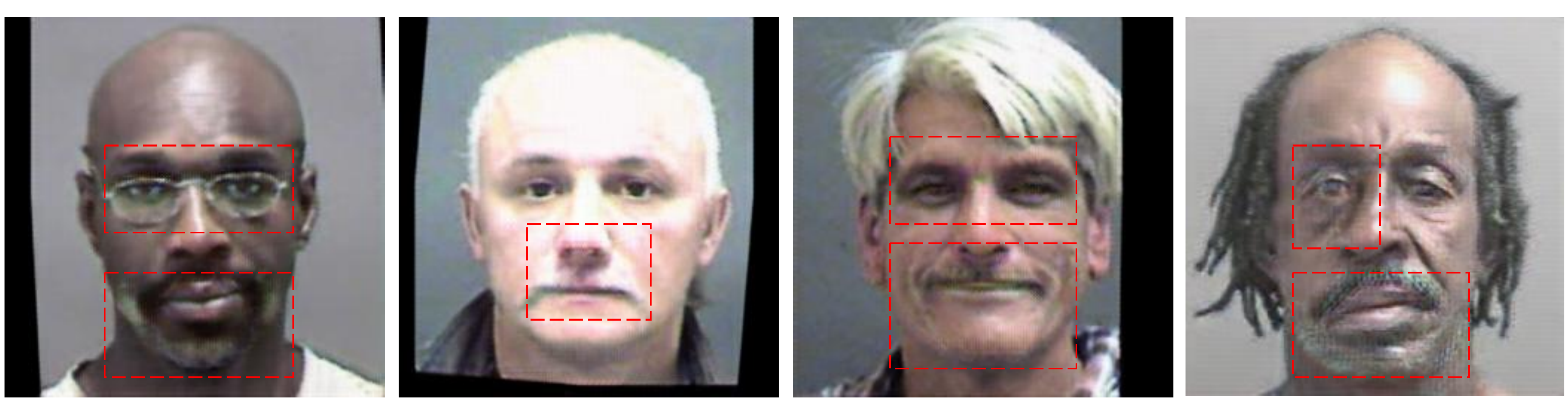}
\caption{Sampled results of basic semantic guided generation (without noise). There are obvious smooth and blurred regions on the faces. There are also artifacts at the junction of different regions. (Both shown in red boxes.) These problem could also be observed in Fig.8[a],[b] and Fig.9[a]. \textbf{Zoom in} for a better view of the details.}

\label{}
\end{figure}

\begin{figure*}[t]
\centering
\includegraphics[width=1.00\linewidth]{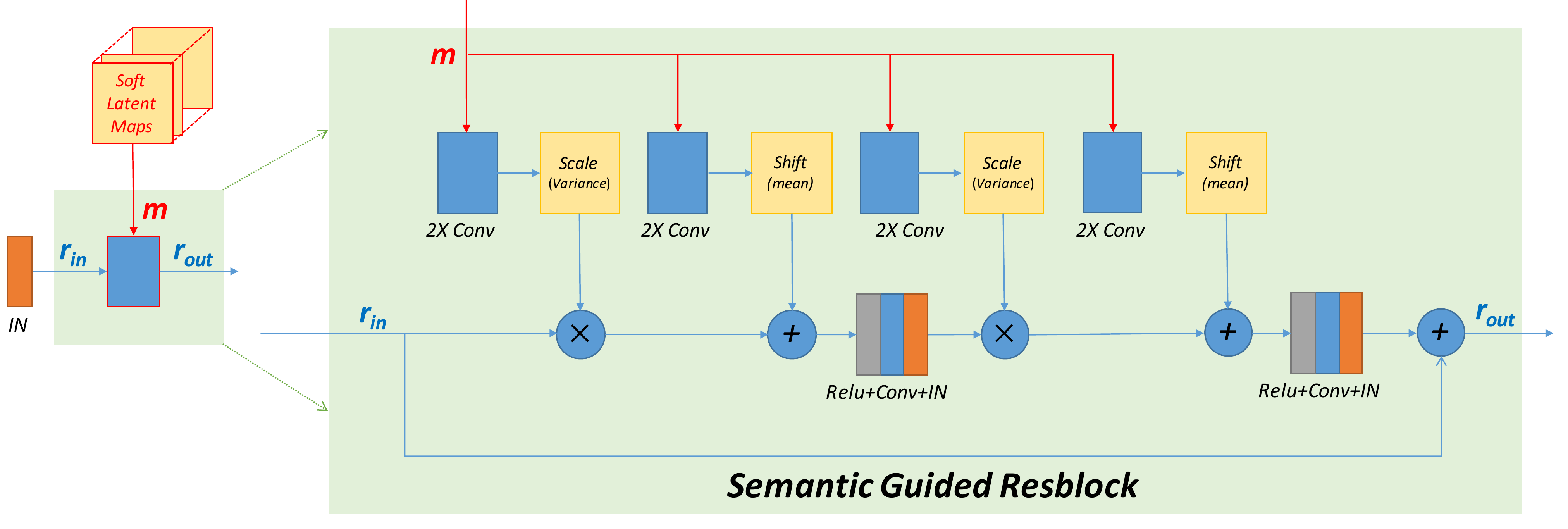}
\caption{The detailed structure of Semantic Guided ResBlocks. In each resblock, ``soft" latent maps are injected twice after the instance normalization (IN) operations. The modulating coefficients (means and variances) are generated by convolutional operations on the ``soft" latent maps respectively.}
\label{fig:comparison}
\end{figure*}

\subsubsection{ProjectionNet}
Privious state-of-the-art method SPT-GAN \cite{17} directly fuses the target age condition with the input face image along the image channels, and then sends them into the encoder.  
The condition is introduced into the network through a ``hard" way, in which the condition is encoded as one-hot encoding maps.
ProjectionNet fuses the low-level semantic layouts with the ``soft" target age condition, and then mixes them into ``soft" latent maps as shown in Fig.3.

There are four steps in ProjectionNet.
Firstly, the parsing layouts are obtained through BiseNet \cite{35} and the corresponding noise map is sampled from the standard Gaussian distribution.
Then, the channel-wise decomposed semantic parts are multiplied with the noise map.
Next, the noisy semantic layouts are concatenated with the ``soft" target age, producing the conditional noisy semantic.
Finally, the conditional noisy semantic maps are projected into ``soft" latent maps through a four-layer convolutional neural networks.
In the ``soft" latent maps, the age condition is integrated into the semantic regions, rather than a separated one-hot encoding.
\subsubsection{ConstraintNet}
ProjectionNet fuses the one-hot condition with the semantic maps integrally. 
However, it does not consider the high-level condition-related spatial context for feature injection.
In image translation tasks, richer spatial information will lead to more detailed face generation through injection operation.
In order to introduce more age-related spatial information into the ``soft" latent maps for the following spatial feature injection, we design an attention-based feature disentanglement architecture.
The spatial information about target age is obtained from the output face, which is constrained and obtained by several strong losses.
Specifically, the downsampled output face is firstly transformed into a group of feature maps $F$ and an attention mask $M$ (constraint to 0-1 by sigmoid operation).
Thus we could obtain two masks, $M1 = M$ and $M2 = (1 - M)$.
Then these two masks are multiplied to the $F$ respectively.
The feature maps $F$ are spatially disentangled into two groups.
To fully disentangle the features, we employ the age annotation (age estimation loss) of the output face to constraint the age-related feature group and use the negative age loss (-$\Delta$ age estimation loss) to constraint the age-irrelevant feature group, in which -$\Delta$ is the weight coefficient of the loss.
Finally, we use the age-related spatial feature maps to constraint the ``soft" latent maps with L1 loss, which endows the ``soft" latent maps with more age-related spatial context. 

\subsubsection{Latent Map Injection}
Different from the previous feature fusion strategy such as concatenation \cite{17}, we apply feature injection in our networks when ``soft" latent maps are obtained.
Inspired by SPADE \cite{26} and CLADE \cite{28}, we employ spatially-adaptive normalization to integrate the age-related context-enhanced ``soft" latent maps into our network.
Concretely, the latent maps are incorporated into the network through Semantic Guided ResBlocks as shown in Fig.5.
For each Semantic Guided ResBlock, combinations of convolutional layers and injection operation are stacked to process the incoming features, and a skip connection is used to preserve information of the input features.
Semantic Guided ResBlock models its output $\mathbf{r}_{out}$ according to the input (denoted as $\mathbf{r}_{in}$) and the latent maps $\mathbf{m}$ learned from the semantic layouts. $\mathbf{r}_{out}$ could be computed as:
\begin{equation}
\mathbf{r}_{out}=\mathbf{r}_{in}\cdot C_{scale}(\mathbf{m}) + C_{shift}(\mathbf{m})
\end{equation}
where $C_{scale}$ and $C_{shift}$ are both two cascaded convolutional layers computing the scaling and shifting parameters.
$\mathbf{r}_{in}$ is the instance normalization output of the previous layer.  
Due to the limitation of the computing resources, we employ the injection operation into the six resblocks between the encoder and decoder with the feature map resolution 64$\times$64 .

\subsubsection{Feature Refinement Module}
Age translation could be regarded as a face manipulation task.
In this task, part of the face structure is manipulated and transformed into a new attribute type, and the rest remains unchanged.
In this paper, we inherit the views in [47] to meet the above characteristics, which the final result is fused from the output of the network and the input image.
In the backbone of the generator, we attach a Feature Refinement Module after the decoder for the final feature fusion and detail refinement.
We employ two types of attention mechanisms, Channel Mixer and Spatial Attention Fusion, which respectively mix the channel-wise information and recombine the spatial context of the decoder output with the input face.

\begin{figure*}[t]
\centering
\includegraphics[width=1.0\linewidth]{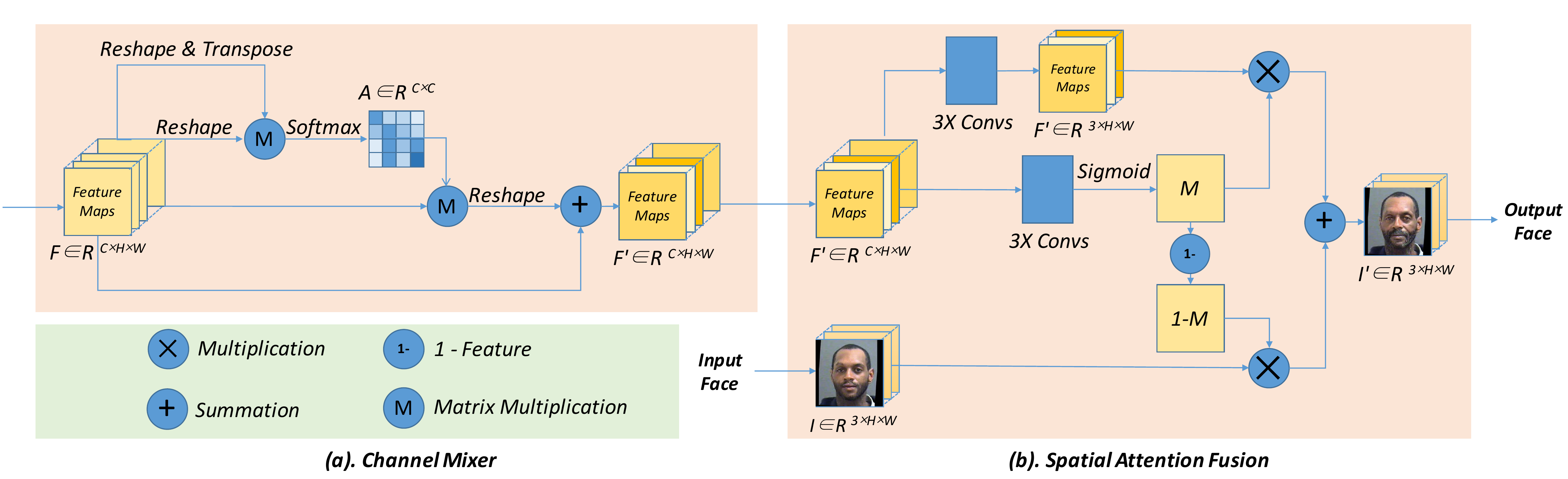}
\caption{The two types of attention mechanism cascaded employed in the Feature Refinement Module. 
(a). Channel Mixer. 
(b). Spatial Attention Fusion.
Channel Mixer is designed to mix channel wise information and improve the effective utilization of channel information.
Spatial Attention Fusion is used to fuse the network output and the original input faces for detail refinement.}
\label{fig6}
\end{figure*}

\emph{\textbf{Channel Mixer.}}
Inspired by DAN \cite{36}, we introduce the self-attention based channel attention to distill channel-wise information in our Feature Refinement Module before spatial fusion.
The Channel Mixer Module is illustrated in Fig.6(a).
In this module, the attention map is calculated from the original feature maps $\textbf{F}\in\mathbb{R}^{C \times H \times W}$.
Firstly, $\textbf{F}$ is reshaped to $\mathbb{R}^{C \times N}$
Then, a matrix multiplication is performed between $\textbf{F}$ and the transpose of $\textbf{F}$ to get the attention map $\textbf{A}\in\mathbb{R}^{C \times C}$.
In addition, we apply a softmax layer to obtain the final normalized attention map:
\begin{equation}
x_{ij}=\frac{exp(F_{i}\cdot F_{j})}{\sum_{i=1}^{C}exp(F_{i}\cdot F_{j})}
\end{equation}
where $x_{ij}$ measures the correlation between the $i^{th}$ channel and $j^{th}$ channel.  
Finally, a matrix multiplication is conducted between the transpose of $\textbf{A}$ and $\textbf{F}$.
Then the output feature maps of the matrix multiplication are reshaped to $\mathbb{R}^{C \times H \times W}$, following an element-wise summation with the identity mapping of  $\textbf{F}$.

\emph{\textbf{Spatial Attention Fusion.}}
Similar to the spatial attention mechanism used in ConstraintNet, we employ the mask-based attention method in Spatial Attention Module for the final feature fusion, instead of the feature decoupling in ConstraintNet.
Through this method, we fuse the features of the generated target face after the Channel Mixer and the input face for the final detail refinement.
The Spatial Attention Fusion Module is illustrated in Fig.6(b).
\subsection{Discriminator}
The discriminator network $D$ in our model serves two goals: measuring the realism of faces and checking the age fulfillment.
Since age changes are mainly reflected in translations of image texture, we adopt the PatchGAN \cite{37} as the discriminator network in our model, which pays more attention to estimating the visual fidelity of local patches.
Meanwhile, we append an auxiliary classification network to the end of the discriminator to predict which age group the input image belongs to.
These functions collaborate to encourage the synthetic images to be indistinguishable from generic ones and present lifelike age signs of the target age group.

\subsection{Objective Functions}
In our architecture, the objective functions exists in four regions, the self-driven training, the condition-driven training, the GAN backbone and the ConstraintNet.
The loss functions in the GAN backbone consist of four parts, 
an adversarial loss, 
an age regression loss, 
an identity perceptual loss, 
and a pixel loss.
The loss functions in ConstraintNet consist of three parts:
an L1 spatial loss,
an age constraint loss,
and an reversed age loss.
The loss functions in the self-driven training consist of two parts,
a supervised reconstruction loss cooperating with the adversarial loss in GAN backbone.
The loss functions in the condition-driven training consist of two parts,
a cycle-consistency loss cooperating with the adversarial loss in GAN backbone.

The adversarial loss describes the objective of a minimax two-player game between the generator $G$ and the discriminator $D$, where it encourages $G$ to fool $D$ with realistic images and penalizes $D$ for failing to distinguish real and fake data samples.
Given an input $X$, we denote the output of the age regressor as $D^{(a)}(X)$ and the output of the realism estimator as $D^{(x)}(X)$.
In this paper, we adopt the least square adversarial loss and the objective could be represented as
\begin{align}
\mathcal{L}_{adv} = &\: \mathbb{E}_{\mathbf{x}_y}[(D^{(x)}_p(G_p(\mathbf{x}_y,\mathbf{y}_{tar_o})) - 1)^2] \nonumber\\
+ &\: \mathbb{E}_{\mathbf{x}_o}[(D^{(x)}_r(G_r(\mathbf{x}_o, \mathbf{y}_{src_y})) - 1)^2] \nonumber\\
+ &\: \mathbb{E}_{\mathbf{x}_o}[(D^{(x)}_p(\mathbf{x}_o) - 1)^2] \nonumber\\
+ &\: \mathbb{E}_{\mathbf{x}_y}[D^{(x)}_p(G_p(\mathbf{x}_y, \mathbf{y}_{tar_o}))^2] \nonumber\\
+ &\: \mathbb{E}_{\mathbf{x}_y}[(D^{(x)}_r(\mathbf{x}_y) - 1)^2] \nonumber\\
+ &\: \mathbb{E}_{\mathbf{x}_o}[D^{(x)}_r(G_r(\mathbf{x}_o, \mathbf{y}_{src_y}))^2]
\label{eq:adversarial_loss}
\end{align}
As mentioned in previous sections, one important issue of age progression and regression is that the generation results should satisfy the target age condition.
In order to enforce the age fulfillment of synthesized images, an age regression loss is adopted, which could be expressed as
\begin{align}
\mathcal{L}_{age\_reg} = &\: \mathbb{E}_{\mathbf{x}_y}[\: \|D_p^{(a)}(G_p(\mathbf{x}_y,\mathbf{y}_{tar_o})) - \mathbf{y}_{tar_o}\|_2 \:] \nonumber \\
                  + &\: \mathbb{E}_{\mathbf{x}_y}[\: \|D_p^{(a)}(\mathbf{x}_y) - \mathbf{y}_{src_y}\|_2 \:] \nonumber \\
                  + &\: \mathbb{E}_{\mathbf{x}_o}[\: \|D_r^{(a)}(G_r(\mathbf{x}_o, \mathbf{y}_{tar_y})) - \mathbf{y}_{tar_y}\|_2 \:] \nonumber \\
                  + &\: \mathbb{E}_{\mathbf{x}_o}[\: \|D_r^{(a)}(\mathbf{x}_o) - \mathbf{y}_{src_o}\|_2 \:]
\label{eq:age_regression_loss}
\end{align}
Optimizing Equation (4) makes the auxiliary regression network $D^{(a)}$ gain the ability of estimating the age of input, which in turn forces $G$ to render accurate age changing effects.

In addition, identity loss ($\mathcal{L}_{id}$) and pixel loss ($\mathcal{L}_{pix}$) are also adopted to maintain consistency in both personalized feature-level and image-level.
To be specific, we utilize the pre-trained LightCNN-29 \cite{39}, denoted as $\phi_{id}$, to extract the identity-related features.
These two loss terms could be formulated as
\begin{align}
\mathcal{L}_{id} = & \: \mathbb{E}_{\mathbf{x}_y}[\: \|\phi_{id}(G_p(\mathbf{x}_y,\mathbf{y}_{tar_o}))-\phi_{id}(\mathbf{x}_y)\|_F^2]\nonumber \\
 + & \: \mathbb{E}_{\mathbf{x}_o}[\|\phi_{id}(G_r(\mathbf{x}_o,\mathbf{y}_{src_y}))-\phi_{id}(\mathbf{x}_o)\|_F^2]
\label{eq:id_loss}
\end{align}
\begin{align}
\mathcal{L}_{pix} = & \: \mathbb{E}_{\mathbf{x}_y}[\| G_p(\mathbf{x}_y,\mathbf{y}_{tar_o})-\mathbf{x}_y\|_F^2]\nonumber \\
 + & \: \mathbb{E}_{\mathbf{x}_o}[\| G_r(\mathbf{x}_o,\mathbf{y}_{src_y})-\mathbf{x}_o\|_F^2]
\label{eq:pix_loss}
\end{align}

Since unpaired data is used in our work, no ground truth supervision is available to regulate the mapping between images of two age groups.
Therefore, a condition-driven cycle-consistency loss is employed to penalize the difference between the input image and its reconstruction (i.e., the result of reverse mapping), which could be formulated as
\begin{align}
\mathcal{L}_{cyc} = &\: \mathbb{E}_{\mathbf{x}_{y}}[\|G_{r}(G_{p}(\mathbf{x}_{y}, \mathbf{y}_ {tar_o}), \mathbf{y}_ {src_y})-\mathbf{x}_{y}\|_{1}] \nonumber\\
+&\: \mathbb{E}_{\mathbf{x}_{o}}[\|G_{p}(G_{r}(\mathbf{x}_{o}, \mathbf{y}_{tar_y}), \mathbf{y}_ {src_o})-\mathbf{x}_{o}\|_{1}]
\label{eq:cyc_loss}
\end{align}

In the self-driven training, the objective function is to reconstruct the input face supervised by itself, which aims at enhance the network ability of conditional generation with ``soft" supervision.
The loss function is listed as
\begin{align}
\mathcal{L}_{self} = & \: \mathbb{E}_{\mathbf{x}_y}[\| G_p(\mathbf{x}_y,\mathbf{y}_{src_y})-\mathbf{x}_y\|_F^2]\nonumber \\
 + & \: \mathbb{E}_{\mathbf{x}_o}[\| G_r(\mathbf{x}_o,\mathbf{y}_{src_o})-\mathbf{x}_o\|_F^2]
\label{eq:pix_loss}
\end{align}

In ConstraintNet, we use two opposite losses to cooperate with the attention mechanism for feature disentanglement, age constraint loss and an reversed age loss.
The age constraint loss is employed to extract age-related features.
The reversed age loss is used to separate age-irrelevant features.
The structure of age estimator $E$ applied on the feature maps $F$ (Age Related $F\_re$, Age Un-Related $F\_un$) is same as the age regressor as $D^{(a)}(I)$.
These two losses could be formulated as
\begin{align}
\mathcal{L}_{age\_est} = 
&\: \mathbb{E}_{\mathbf{x}_y}
[\: \|E(F\_re) - \mathbf{y}_{tar_o}\|_2 \:] \nonumber \\
- \Delta &\: \mathbb{E}_{\mathbf{x}_y}
[\: \|E(F\_un) - \mathbf{y}_{tar_o}\|_2 \:] \nonumber \\
+ &\: \mathbb{E}_{\mathbf{x}_o}
[\: \|E(F\_re) - \mathbf{y}_{tar_y}\|_2 \:] \nonumber \\
- \Delta &\: \mathbb{E}_{\mathbf{x}_o}
[\: \|E(F\_un) - \mathbf{y}_{tar_y}\|_2 \:]
\label{eq:age_regression_loss}
\end{align}

The primary objective of ConstraintNet is providing age-related features $F\_re$ to endow the ``soft" latent maps $F\_soft$ in ProjectionNet with more age-related spatial context.
So the constraint loss is:
\begin{align}
\mathcal{L}_{constraint} = & \: \mathbb{E}_{\mathbf{x}_y}[\|F\_re - F\_soft \|_F^1]\nonumber \\
+ & \: \mathbb{E}_{\mathbf{x}_o}[\|F\_re - F\_soft \|_F^1]
\label{eq:age_regression_loss}
\end{align}

The overall loss of the proposed method could be concluded as the combination of all previously defined losses:
\begin{align}
\mathcal{L}_{total} = 
&\mathcal{L}_{adv} + 
\lambda_{id}\mathcal{L}_{id} + 
\lambda_{pix}\mathcal{L}_{pix}\nonumber \\
&+
\lambda_{cyc}\mathcal{L}_{cyc} + \lambda_{self}\mathcal{L}_{self}\nonumber \\
&+
\lambda_{age\_reg}\mathcal{L}_{age\_reg} +
\lambda_{age\_est}\mathcal{L}_{age\_est}\nonumber \\
&+
\lambda_{constraint}\mathcal{L}_{constraint}
\label{eq:total_loss}
\end{align}

where  $\lambda_{id}$, $\lambda_{pix}$, $\lambda_{cyc}$, $\lambda_{self}$, $\lambda_{age\_reg}$, $\lambda_{age\_est}$,  and $\lambda_{constraint}$ are coefficients balancing the relative importance of each loss term.
Finally, $G_p$, $G_r$, $D_p$, and $D_r$ are solved by optimizing:
\begin{equation}
\min_{G_p, G_r} \max_{D_p, D_r} \mathcal{L}_{total}
\end{equation}

For better illustration of the overall procedure of the Noisy-semantic guided GANs, the pseudo code is shown in \textbf{Algorithm 1}.

\begin{algorithm}[t]
\caption{The pseudo code of NSG-GAN.}
\label{algorithm1}
\begin{algorithmic}[1]
\Require
~~\\
$\mathbf{x}_y$: Image data young face

$\mathbf{y}_{tar\_o}$: Target age of face $\mathbf{x}_y$

$\mathbf{y}_{src\_y}$: Source age of face $\mathbf{x}_y$\\
$\mathbf{x}_o$: Image data old face;

$\mathbf{y}_{tar\_y}$: Target age of face $\mathbf{x}_o$

$\mathbf{y}_{src\_o}$: Source age of face $\mathbf{x}_o$\\

$E$: The number of epoch $=$ 40;\\
$lr$: Learning rate;
\Ensure
~~\\
$\mathbf{x}_{y\rightarrow o}$: The prediction from $\mathbf{x}_y$;\\
$\mathbf{x}_{o\rightarrow y}$: The prediction from $\mathbf{x}_o$;\\
$\mathbf{x}_o'$: The reconstruction of $\mathbf{x}_o$;\\
$\mathbf{x}_y'$: The reconstruction of $\mathbf{x}_y$;\\

\textbf{Step1:} Image pre-processing;

\State Align the faces using MTCNN;

\State Resized into 256 $\times$ 256;\\

\textbf{Step2:} Network Initialization;

\State Initialize weights of $G_p$, $G_r$, $D_p$, $D_r$, 

$w = \{ w_{G_p},w_{G_r},w_{D_p},w_{D_p} \} = 0$;\\

\textbf{Step3:} Network Training;
\For{$E = 0$ to 40 :}

\textbf{Condition-Driven Training:}

Compute the prediction: $\mathbf{x}_{y\rightarrow o} = G_p(\mathbf{x}_y,\mathbf{y}_{tar_o})$;

Compute the cycle reconstruction:

$\mathbf{x}_y'= G_{r}(G_{p}(\mathbf{x}_{y}, \mathbf{y}_ {tar_o}), \mathbf{y}_ {src_y})$

Compute the prediction: $\mathbf{x}_{o\rightarrow y} = G_r(\mathbf{x}_o,\mathbf{y}_{tar_y})$;

Compute the cycle reconstruction:

$\mathbf{x}_o'= G_{p}(G_{r}(\mathbf{x}_{o}, \mathbf{y}_ {tar_y}), \mathbf{y}_{src_o})$

\textbf{Self-Driven Training:}

Compute the self reconstruction: 

$\mathbf{x}_y' = G_p(\mathbf{x}_y,\mathbf{y}_{src_y})$;

Compute the self reconstruction: 

$\mathbf{x}_o' = G_r(\mathbf{x}_o,\mathbf{y}_{src_o})$;

\textbf{Weights Updating:}

Compute the loss $L = \min_{G_p, G_r} \max_{D_p, D_r} \mathcal{L}_{total}$;

Compute gradient $g =\bigtriangledown L$;

Update the weights $w = w - lr \times g$;

\EndFor

\State Compute the final prediction with new $\mathbf{x}_y, \mathbf{x}_o$: 

$\mathbf{x}_{y\rightarrow o} = G_p(\mathbf{x}_y,\mathbf{y}_{tar_o})$.

$\mathbf{x}_{o\rightarrow y} = G_r(\mathbf{x}_o,\mathbf{y}_{tar_y})$
\end{algorithmic}
\end{algorithm}

\begin{table*}[t]
\centering
\caption{Results on Age Estimation Error and Identity Verification Rate on MORPH and CACD. Mean verification scores are shown in brackets after the total verification rate.}
\begin{tabular}{c|c|c|c|c|c}
\hline
\multicolumn{2}{c|}{} & \multicolumn{2}{c|}{\textbf{MORPH}}                        & \multicolumn{2}{c}{\textbf{CACD}}                         \\ \hline
Method        & Year   & Age Estimation Error & Identity Verification Rate & Age Estimation Error & Identity Verification Rate \\ \hline
CAAE          & 2017   & 10.34 ± 5.63         & 34.83 (71.75)              & 5.16 ± 7.08          & 3.59 (59.90)               \\ \hline
IPC-GAN       & 2018   & 1.74 ± 7.44          & 99.86 (94.04)              & 8.11 ± 9.69          & 99.19 (91.60)              \\ \hline
Dual cGAN    & 2018   & 2.44 ± 6.03          & 99.99 (93.15)              & 3.28 ± 8.01          & 99.88 (93.85)              \\ \hline
SPT-GAN       & 2020   & 1.53 ± 6.50          & 100.00 (95.67)             & 1.78 ± 7.53          & 99.92 (96.13)              \\ \hline
NSG-GAN       & 2021   & \textbf{1.20 ± 6.81}          & 99.99 (95.27)              & \textbf{1.45 ± 8.02}          & \textbf{99.93 (94.20)}              \\ \hline
\end{tabular}
\end{table*}

\section{Experiments}
\subsection{Dataset and Implementation Details}
In this paper, we mainly conduct experiments on publicly available datasets, MORPH \cite{40} and CACD \cite{41}, which are widely used for evaluating face age translation algorithms.
MORPH contains 55,349 images from 13,672 people with age ranges from 16 to 77 years old.
Images in this dataset are captured in constrained environment and thus are of high quality.
Age annotations are also provided along with the image data.
CACD contains 163,346 images from 2,000 celebrities at different ages.
These images are collected from the Internet via Google Image Search, containing large variations in pose, illumination and expression.
Moreover, non-face images and incorrect age labels make CACD a dataset much more challenging than MORPH.
Though there are other age related datasets such as FG-NET \cite{42} and UTKFace \cite{43}, we do not use them in this paper for the following two reasons.
(1) To maintain a consistent comparison with the previous unified methods in Table.I.
(2)The size of these two datasets are limited, and they cannot get high-quality semantic parsing using the open source methods.

Following previous works \cite{17,44}, we divide images in MORPH and CACD into four age groups (30-, 31-40, 41-50, 51+).
For each dataset, we randomly choose $80\%$ images for training and the rest $20\%$ images for testing.
Images in both MORPH and CACD are aligned by the location of both eyes and then resized to $256\times 256$ before fed into the network.
All face images in MORPH and CACD are aligned using MTCNN \cite{45}. 
Noisy images and some wrong faces are artificially filtered out.
For semantic layouts, we use a pre-trained face parsing network BiseNet \cite{35} to acquire the segmentation results.
And in order to prevent semantic ambiguity, we merge all facial parts into 12 classes which are closely related to different face structural regions, two eyes, two eyebrows, two ears, glasses, upper and lower lips, inner mouth, hair, nose, skin, neck, cloth, and background.

At the training stage, we select Adam ($\beta_1 = 0.5$, $\beta_2 = 0.999$) as the optimizer, with learning rate and batchsize set to $1\times 10^{-4}$ and 24 respectively.
The balancing parameters $\lambda_{id}$, $\lambda_{pix}$, $\lambda_{cyc}$, $\lambda_{self}$, $\lambda_{age\_reg}$, $\lambda_{age\_est}$,  and $\lambda_{constraint}$ are set to 10, 1, 1, 10, 800, 10, 1, and in $\lambda_{age\_est}$, $\Delta$ is set to 0.5 both for MORPH and CACD.
Our model is trained on MORPH for 40 epochs and on CACD for 35 epochs.

\subsection{Evaluation Metrics}
Quantitative experiments are conducted to provide an objective measurement of our model performance and the comparisons against benchmarks.
As for the evaluation metrics, \textbf{Age translation accuracy} and \textbf{identity preservation rate} are two essential indicators for the biphasic facial age translation.
They measure how well the model fulfill the age translation while maintaining the identity information in the input image.
To ensure the fairness of assessment, public available face analysis tools, i.e., Face++ APIs [46], are used in our experiments.

\subsubsection{Age Translation Accuracy}
The fundamental goal of biphasic facial age translation is to render age changing effects on input face images.
Therefore, it is critical for the generated images fall exactly into the correct age group.
To this end, we estimate the age of each synthesized image and then compare statistics of distribution (\textbf{absolute error of mean ages}) for generic and synthetic images.
To evaluate the performance of the proposed method objectively, the measurement of age estimation is conducted via the public face analysis API of Face++ for both real and fake data.

\begin{table*}[]
\centering
\caption{The specific prediction results of estimated ages and errors for each age group.}
\begin{tabular}{c|c|c|c|c|c|c|c|c}
\hline
               & \multicolumn{4}{c|}{\textbf{MORPH}}                                & \multicolumn{4}{c}{\textbf{CACD}}                                   \\ \hline
Age Group      & 30 -         & 31 - 40      & 41 - 50      & 51 +         & 30 -         & 31 - 40      & 41 - 50       & 51 +          \\ \hline
Generic        & 27.80 ± 5.60 & 38.60 ± 7.43 & 47.74 ± 8.30 & 57.25 ± 8.29 & 25.55 ± 9.02 & 38.50 ± 9.66 & 48.53 ± 10.69 & 53.41 ± 12.41 \\ \hline
NSG-GAN        & 29.12 ± 6.91 & 37.35 ± 7.03 & 46.63 ± 6.15 & 56.13 ± 7.16 & 26.50 ± 6.95 & 36.27 ± 8.57 & 49.98 ± 7.84  & 54.58 ± 9.12  \\ \hline
Absolute Error & +1.32        & -1.25        & -1.11        & -1.12        & +0.95        & -2.23        & +1.45         & +1.17         \\ \hline
\end{tabular}
\end{table*}

\begin{figure*}[t]
\centering
\includegraphics[width=1.00\linewidth]{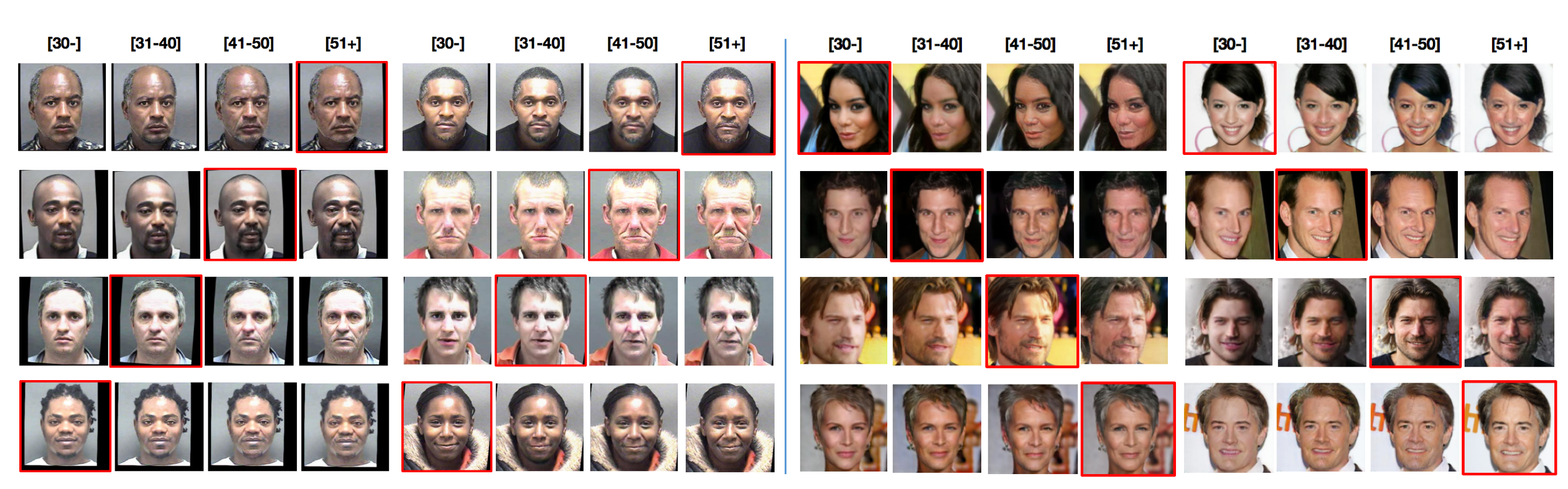}
\caption{Randomly sampled results generated by NSG-GAN on MORPH (left eight columns) and CACD (right eight columns) datasets. Red box indicates the input face image for each age progression/regression sequence. Zoom in for a better view of the details.}
\label{fig:comparison}
\end{figure*}

\subsubsection{Identity Preservation Rate}

While input face images should be manipulated to present desired age changing effects, the identity information embedded should remain intact.
To this end, face verification rate between input and output is computed to check whether identity permanence is achieved in generated images.
The threshold is set to 73.975@FAR=1e-5 for our identity preservation experiments using Face++ APIs.

\subsection{Quantitative Results}
\subsubsection{Age Translation Accuracy}

In this paper, we estimate the age of each synthesized image and then compare the statistics of distribution  between the generic and generated faces. The absolute error of mean ages is the the most important indicator.
According to results reported in Table.I, our method achieves the smallest absolute error in mean ages and outperforms other biphasic approaches by a clear margin.
(In all Tables, the noisy-semantic guided GANs is abbreviated as NSG-GAN.)
Signs of aging are heavily blurred in results produced by CAAE \cite{2}, leading to large error in estimated ages.
IPC-GAN \cite{15} achieves satisfying results on MORPH but much worse on CACD, indicating its incapability in dealing with in-the-wild images.
However, with the prior knowledge provided with the parsing map, our model could understand the semantic layout of the input image without being interfered by complex textures, resulting in an outstanding performance on MORPH and CACD.
Our age estimation error is lower than previous state-of-the-art biphasic facial age translation method SPT-GAN\cite{17} in both datasets.
SPT-GAN just introduces spatial attention to restrain age unrelated features, which have not paid attention to the characteristic changes of specific semantic regions and not fully explored channel and spatial information.
The specific prediction results and errors of NSG-GAN for all age groups are shown in TABLE.II.
The Generic indicates the original real faces in these two datasets. The NSG-GAN means the generated faces through our method.
Though there are still some gaps between the indicators of NSG-GAN and the one-direction algorithms such as A3GAN \cite{45}, the indicators of NSG-GAN have reached the state-of-the-art performance for biphasic facial age translation.
\subsubsection{Identity Preservation Rate}
As shown in TABLE.I, our method achieves the identity preservation rate of over $99\%$ on both two datasets, which is comparative to other approaches, indicating the effectiveness of our method in preserving personal characteristics of input faces.
Our NSG-GAN achieves $99.99\%$ for MORPH dataset with an average verificaiton score $95.27$ and $99.93\%$ for CACD dataset with an average verificaiton score $94.20$.
All the metrics are satisfactory for the cross-age face recognition system.

\subsection{Qualitative Results}
Randomly sampled results on MORPH and CACD are shown in Fig. 7.
Although input images cover a widely range of age, gender, pose, and expression, visually plausible face images presenting the target age could be synthesized by our model.
Moreover, using parsing layouts as guidance enhances the model to learn the semantic-region transformations of input face images, which enables fine-grained translations with more natural age changes.
Meanwhile, the semantic layouts could also inform the model with the spatial distribution of each facial component, resulting in reasonable changes of facial texture.
Different from the previous methods, the performance of NSG-GAN shows the gradual change in the adjacent age groups without exaggerated facial changes, which is more in line with the real situation.

\subsection{Ablation Analysis}
In this section, we conduct several ablation studies on MORPH dataset to evaluate the contributions of the proposed components to the overall performance of our model.
In this part, we utilize the Age Translation Accuracy and Identity Verification Rate as our evaluation metrics.
Specifically, we perform the following experiments for overall analysis of the effectiveness of our method.
(1). Effect of Module Ablation.
(2). Effect of Noisy-Semantic Injection.
(3). Effect of Noise Integration.
(4). Effect of Age-related Feature Disentanglement.
(5). Effect of Joint Training Strategy.

\begin{table*}[t]
\centering
\caption{The ablation analysis of different modules.}
\begin{tabular}{c|c|c|c|c|c|l|l|l|l}
\hline
\multicolumn{4}{c|}{Module Ablatian}                                                                                                                              & \multicolumn{6}{c}{Metrics}                                                                                                                            \\ \hline
\begin{tabular}[c]{@{}c@{}}Encoder-decoder\\ (Baseline)\end{tabular} & ProjectionNet & ConstraintNet & \begin{tabular}[c]{@{}c@{}}Feature Refinement \\ Module\end{tabular} & \begin{tabular}[c]{@{}c@{}}Identity Verification \\ Rate\end{tabular} & \multicolumn{5}{c}{\begin{tabular}[c]{@{}c@{}}Age Translation \\ Accuracy\end{tabular}} \\ \hline
\checkmark                                                                    &               &               &                                                                      & 98.96 (90.14)                                                         & \multicolumn{5}{c}{4.07 ± 7.38}                     \\ \hline
\checkmark                                                                    & \checkmark             &               &                                                                      & 99.13 (94.14)                                                         & \multicolumn{5}{c}{1.97 ± 7.56}                     \\ \hline
\checkmark                                                                    & \checkmark             & \checkmark             &                                                                      & 99.97 (95.16)                                                         & \multicolumn{5}{c}{1.39 ± 8.18}        \\ \hline
\checkmark                                                                    & \checkmark             & \checkmark             & \checkmark                                                                    & \textbf{99.99 (95.27)}                                                         & \multicolumn{5}{c}{\textbf{1.20 ± 6.81}}                                                         \\ \hline
\end{tabular}
\end{table*}

\subsubsection{Module Ablation}

There are four modules in the overall NSG-GAN framework, which are encoder-decoder, ProjectionNet, ConstraintNet, and Feature Refinement Module.
The encoder-decoder is the baseline method with two downsampling layers, six resblocks, and two unsampling layers, which is the basic architecture inherited from CycleGAN.
The ProjectionNet is to introduce the low-level structural semantic information for injection, which fuses the condition and noisy semantic layouts into ``soft" latent maps with the design in Fig.3.
The ConstraintNet is to constraint the ``soft" latent maps with high-level spatial age-related features.
The Feature Refinement Module is to refine the final details.
The ablation results are shown in TABLE.III.
We could find that ProjectionNet and ConstraintNet bear the most important contributions.
Each module has its own role for the final performance.
The final NSG-GAN achieves the best results.
\subsubsection{Noisy-Semantic Injection}
In ProjectionNet, we compare four architectures to find the most effective network structure for semantic injection, which are (1) Semantic Only, (2) Conditional Semantic, (3) Conditional Noisy Semantic, (4) Conditional Noisy Semantic without eyes and lips.
Semantic Only means that the ProjectionNet only projects the semantic into latent feature maps. 
The target age (condition) is delivered into the encoder with the input face.
The Conditional Semantic indicates that we combine the semantic layouts with the target age.
Then the conditional semantic layouts are mapped to latent maps.
Through this comparison, we aim to find the better branch for condition fusion, the projection branch or the backbone branch.
Conditional Noisy Semantic means the sampled noisy layout is multiplied on the semantic maps.  
Meanwhile, we combine the noisy semantic layouts with the one-hot target age.
In this case, the purpose of adding noise is to increase the randomness of the face area and avoid smoothing.
The detailed analysis of noise is presented in the following subsection.
The comparisons of Semantic Injection are shown in TABLE.IV.

\begin{table}[]
\centering
\caption{The ablation analysis of different semantic injection types.}
\begin{tabular}{c|c|c}
\hline
\begin{tabular}[c]{@{}c@{}}Semantic Injection \\ Types\end{tabular}                        & \begin{tabular}[c]{@{}c@{}}Identity Verification \\ Rate\end{tabular} & \begin{tabular}[c]{@{}c@{}}Age Translation \\ Accuracy\end{tabular} \\ \hline
Semantic Only                                                                              & 98.52 (92.72)                                                         & 3.20 ± 6.29                                                         \\ \hline
\begin{tabular}[c]{@{}c@{}}Conditional \\ Semantic\end{tabular}                                                                       & 98.92 (95.27)                                                         & 1.58 ± 7.11                                                         \\ \hline
\begin{tabular}[c]{@{}c@{}}Conditional \\ Noisy Semantic\end{tabular}                      & \textbf{99.99 (96.14)}                                                         & \textbf{1.18 ± 7.82}                                                         \\ \hline
\begin{tabular}[c]{@{}c@{}}Conditional \\ Noisy Semantic\\ (No eyes and lips)\end{tabular} & 99.99 (95.27)                                                         & 1.20 ± 6.81                                                       \\ \hline
\end{tabular}
\end{table}

In order to make more in-depth use of the semantic, and inspired by fuzzy small areas of the previous results shown in Fig.8[a]-[b], we explore to use local semantic for injection, removing the eye area and lip area shown in ProjectionNet of Fig.3.
In this way, the small areas are transformed based on the encoder-decoder structure, ignoring the injection operations.
Meanwhile, the characteristics of age change are hardly reflected in the eyes and lips.
Though the metrics in TABLE.IV show that the Conditional Noisy Semantic with eyes and lips are better in quantitative analysis, they are not satisfactory for visual quality. 
The clear eyes and lips generated by the semantic without small areas can improve the fidelity of image observation as shown in Fig.8.
In Fig.8. we could find that the column [a] and column [b] are relatively blurred, the column [c] and column [d] are more clear.
Moreover, [d] has clearer lips and more vivid eyes than [c] as shown in the red boxes between [c] and [d].
This brings more realism to face generation.
Therefore, we choose the last design, Conditional Noisy Semantic without eyes and lips as our final architecture.
\begin{figure}[]
\centering
\includegraphics[width=0.9\linewidth]{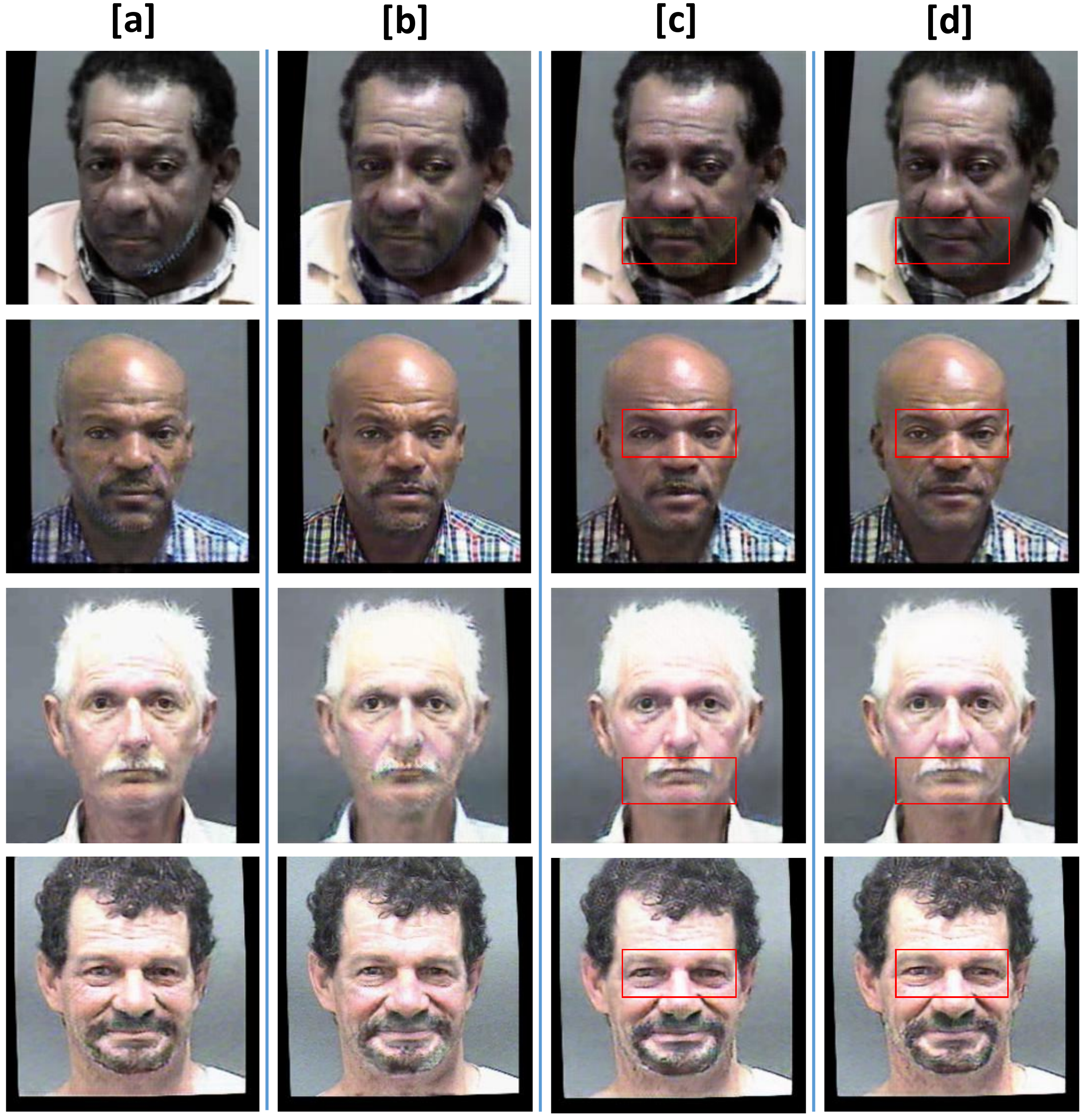}
\caption{Sampled results of different Semantic Injection types. These four-column faces are generated from [a] Semantic Only, [b] Conditional Semantic, [c] Conditional Noisy Semantic, [d] Conditional Noisy Semantic without eyes and lips. 
Each row includes faces of the same age, generated by the four methods. 
Red box indicates the details in the eye and lip regions. \textbf{Zoom in} for a better view of the details.}
\label{fig1}
\end{figure}

\subsubsection{Noise Integration}
In this section, we verify the role of noise in the network.
We introduce the noises into two places of our NSG-GAN, the ProjectionNet and the decoder.
As aforementioned, the purpose of adding noise in ProjectionNet is to increase the randomness of the face area and avoid smoothing.
The noise in the decoder is to refine the details after the upsampling layers with the size of 128 $\times$ 128 and 256 $\times$ 256.
The effect of noises can be demonstrated through the indicators of TABLE.V, while the the visualization results can better reflect their significance as shown in Fig.9.
We could find that the noisy semantic prevents large areas of the face from being smooth and blurred as the column [a], which could result in unreality.
And the noise in the decoder improves the quality of the details compared with column [b]. 
This advantage is also shown in Fig.8, in which the fidelity of [c] and [d] is better than the blurred [a] and [b].

\begin{table}[]
\centering
\caption{The ablation analysis of noise integration.}
\begin{tabular}{c|c|c|c}
\hline
\multicolumn{2}{c|}{Noise Integration Positions} & \multirow{2}{*}{\begin{tabular}[c]{@{}c@{}}Identity Verification \\ Rate\end{tabular}} & \multirow{2}{*}{\begin{tabular}[c]{@{}c@{}}Age Translation \\ Accuracy\end{tabular}} \\ \cline{1-2}
ProjectionNet           & Decoder             &                                                                                           &                                                                                         \\ \hline
                        &                     &98.53 (93.22)                                                            & 1.95 ± 6.87                                                                             \\ \hline
\checkmark               &                     & 99.99 (94.68)                                                                             & 1.25 ± 8.26                                                                             \\ \hline
\checkmark               & \checkmark           & \textbf{99.99 (95.27)}                                                                             & \textbf{1.20 ± 6.81}                                                                            \\ \hline
\end{tabular}
\end{table}

\begin{figure}[]
\centering
\includegraphics[width=0.75\linewidth]{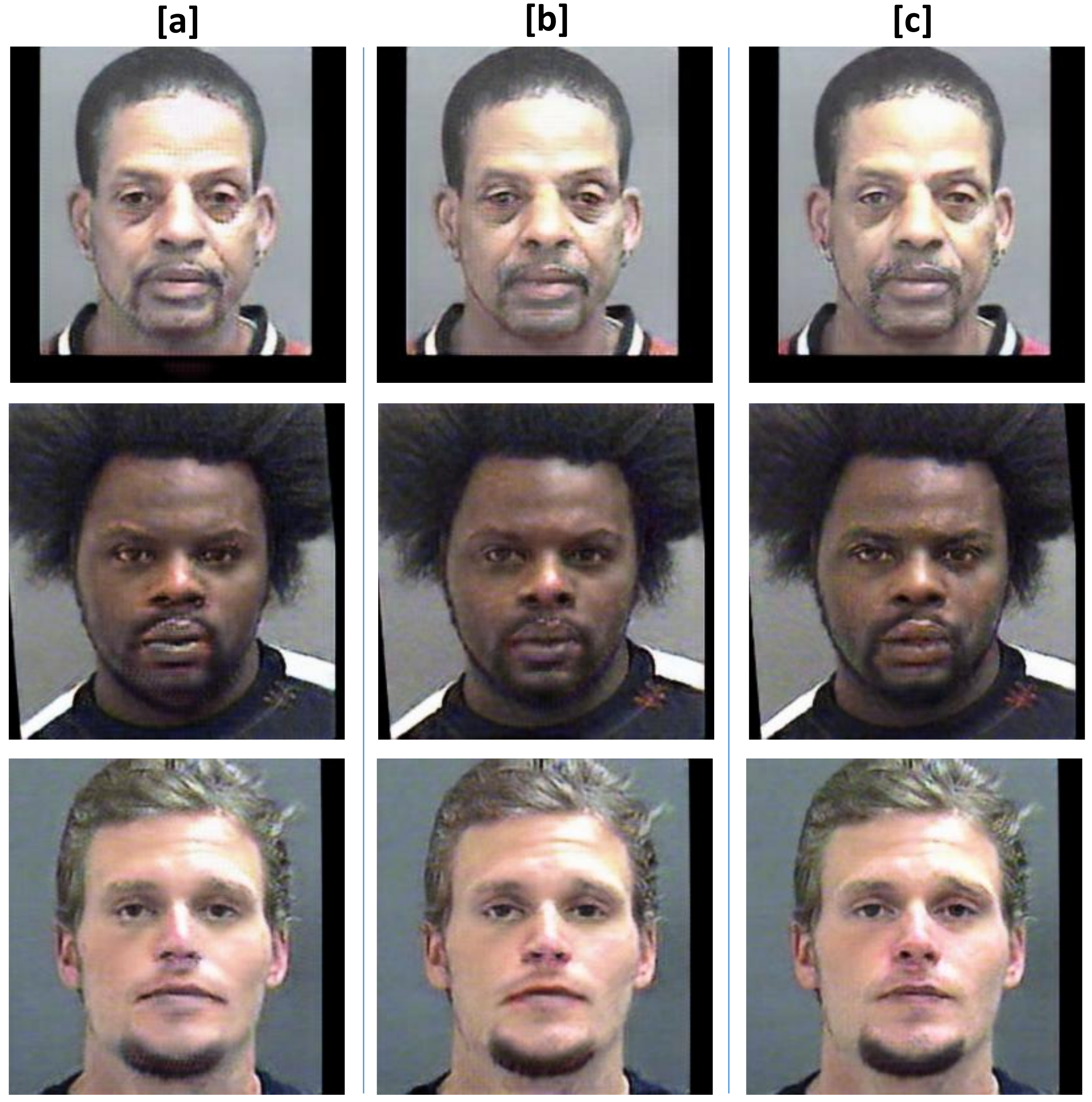}
\caption{Sampled results of different Noise Integration Strategies. These three-column faces are generated from [a] No Noise, [b] Add Noise in ProjectionNet, [c] Add Noise in ProjectionNet and Decoder.
Each row includes faces of the same age generated by the three strategies. 
\textbf{Zoom in} for a better view of the details.}
\label{fig9}
\end{figure}

\subsubsection{Age-related Feature Disentanglement}
In ConstraintNet, we compare three types of constraints for the ``soft" latent maps.
Our original intention to add this network is to provide the transformed facial spatial context for the feature map of the injection operation.
The first architecture is using a simple mapping network which directly transfers the output face to constrain the ``soft" latent maps.
The second architecture is to employ the attention-based feature disentanglement network, in which we use the age loss for age-related feature constraint and identity perceptual loss for age-irrelevant feature constraint. 
The third architecture is also to employ the attention-based feature disentanglement network, which utilizes the $-\Delta$ age loss for age-irrelevant feature constraint.
The results are illustrated in TABLE.VI.
We could find that the feature disentanglement network provides enough spatial information for age translation.
The simple mapping network has a negative effect on the transmission of spatial information, because there are no specific constraints.
And different from other previous studies\cite{34}, in which identity loss is used to constrain attribute independent feature, we use $-\Delta$ age loss for age-irrelevant feature constraint.
Through this loss, the performance of our age translation is better, 

\begin{table}[]
\centering
\caption{The ablation analysis of Age-related Feature  Disentanglement.}
\begin{tabular}{c|l|c|c}
\hline
\multicolumn{2}{c|}{\multirow{2}{*}{\begin{tabular}[c]{@{}c@{}}ConstraintNet\\ Types\end{tabular}}}           & \multirow{2}{*}{\begin{tabular}[c]{@{}c@{}}Identity Verification \\ Rate\end{tabular}} & \multirow{2}{*}{\begin{tabular}[c]{@{}c@{}}Age Translation \\ Accuracy\end{tabular}} \\
\multicolumn{2}{c|}{}                                                                                         &                                                                                        &                                                                                      \\ \hline
\multicolumn{2}{c|}{Simple Mapping}                                                                           & 98.34 (93.80)                                                                          & 8.20 ± 7.29                                                                          \\ \hline
\multicolumn{2}{c|}{\begin{tabular}[c]{@{}c@{}}Feature Disentanglement\\ (With Identity Loss)\end{tabular}}   & 99.99 (96.10)                                                                          & 2.08 ± 6.49                                                                          \\ \hline
\multicolumn{2}{c|}{\begin{tabular}[c]{@{}c@{}}Feature Disentanglement\\ (With $-\Delta$ Age Loss)\end{tabular}} & 99.99 (95.27)                                                                          & \textbf{1.20 ± 6.81}                                                                          \\ \hline
\end{tabular}
\end{table}

\subsubsection{Joint Training Strategy}
Due to the inherent disadvantage of lacking paired data in the biphasic facial age translation task and inspired by MUSTGAN\cite{19}, we introduce three types of learning strategies jointly used in this paper,
(1) unsupervised condition-driven cycle-consistent generation (basic conditional CycleGAN training strategy),
(2) supervised self-driven generation,
(3) mixed training of (1) and (2).
The results is shown in TABLE.VII.
In this section, we use the Age Translation Accuracy to evaluate the network performance.

\begin{table}[h]
\centering
\caption{The ablation analysis of Training Strategy.}
\begin{tabular}{c|l|c}
\hline
\multicolumn{2}{c|}{\multirow{2}{*}{Training Strategy}} & \multirow{2}{*}{\begin{tabular}[c]{@{}c@{}}Age Translation Accuracy\end{tabular}} \\
\multicolumn{2}{c|}{}                                   &                                                                                      \\ \hline
\multicolumn{2}{c|}{Self-Driven Only}                   & 4.35 ± 6.80                                                                          \\ \hline
\multicolumn{2}{c|}{Condition-Driven Only}              & 2.28 ± 7.98                                                                          \\ \hline
\multicolumn{2}{c|}{Jointly Strategy}                   & \textbf{1.20 ± 6.81}                                                                          \\ \hline
\end{tabular}
\end{table}

Different from the human generation task in MUSTGAN, using self-driven training merely in this paper can not reach the level of condition-driven training.
This is because the face age generation task is not structural enough as the human generation task which relies on the body skeleton.
The age-related feature could not be distinctly decoupled as the skeletons.
However, the supervised self-driven training could promote the age translation task when working with the condition-driven training as shown in TABLE.VII.
At last, we choose the jointly training strategy for NSG-GAN.

\section{Discussion and Conclusion}
\subsection{Limitation and Discussion}
In this paper, we propose noisy-semantic guided GANs, a GAN-based method to solve biphasic facial age translation in a unified way.
Different from the previous methods which use fusion strategy to import the target condition into the backbone network, we attempt to inject the condition with the face semantic layouts into the networks.
We regard this task as an image transfer task between different ages.
In this way, the age translation is conducted on different facial parts, which brings more detailed context.

Although some improvements have been made in the indicators, this method still has its inherent defects.
Firstly, the NSG-GAN relies heavily on high-quality semantic segmentation masks.
Some low-quality segmentation masks will lead to wrong image conversion. 
For example, some faces with glasses will produce artifacts around the eyes.
Secondly, the unified model is complicated which requires high computational costs and relies on a large number of parameter adjustment experiments.
In our experiments, we utilize 8 NVIDIA TITAN XP GPUs to train our networks with the batch size of 16.
This is a disadvantage compared with the previous one-direction method.
Thirdly, the NSG-GAN pays more attention to the texture changes of the face and lacks the treatment of hair.

Furthermore, we also have several suggestions for age translation tasks after this work.
The age-related datasets are relatively large, and the annotations of some images are not accurate, such as CACD.
As a human face manipulation task, does age translation really need such a large coarse dataset?
A dataset with a certain number of high-quality various face images of different ages should be enough for dealing with this task.
Meanwhile, we should use a consistent age estimation method in the process of image annotation and algorithm evaluation.
Recently, researchers generally use public open source tools for evaluation such as the Face++ API.
This is different from the annotation method referenced by model training.
With the update of open source tools online, the comparison of experimental results may bring errors.
Therefore, more fair evaluation metrics need to be developed in the future.
\subsection{Conclusion}
In this paper, we propose NSG-GAN, a unified Noisy Semantic Guided Generative Adversarial Network for biphasic facial age translation.  
Specifically, we propose two novel sub-networks to introduce low-level structural semantic information and high-level age-related spatial features for refined feature injection. 
Meanwhile, in order to mine the strongest mapping ability of the network, we embed two types of learning strategies in the training procedure, supervised self-driven generation and unsupervised condition-driven cycle-consistent generation.
Our method achieves state-of-the-art performance in the biphasic facial age translation task.
In the future, we will promote this method, pay more attention to the dataset analysis, and try to explore more comprehensive and fair evaluation indicators.



\ifCLASSOPTIONcaptionsoff
  \newpage
\fi

\end{document}